\documentclass{article}

\usepackage{microtype}
\usepackage{graphicx}
\usepackage{subfigure}
\usepackage{booktabs} 

\usepackage{hyperref}

\usepackage[accepted]{icml2021}

\icmltitlerunning{A comparative study of stochastic and deep generative models for multisite precipitation synthesis}

\begin{document}

\twocolumn[
\icmltitle{A comparative study of stochastic and deep generative models for multisite precipitation synthesis}



\icmlsetsymbol{equal}{*}

\begin{icmlauthorlist}
\icmlauthor{Jorge Guevara}{to}
\icmlauthor{Dario Borges}{to}
\icmlauthor{Campbell Watson}{to}
\icmlauthor{Bianca Zadrozny}{to}
\end{icmlauthorlist}

\icmlaffiliation{to}{IBM Research}

\icmlcorrespondingauthor{Jorge Guevara}{jorgegd@br.ibm.com}

\icmlkeywords{Machine Learning, ICML}

\vskip 0.3in
]



\printAffiliationsAndNotice{}  

\begin{abstract}
Future climate change scenarios are usually hypothesized using simulations from weather generators. However, there only a few works comparing and evaluating promising deep learning models for weather generation against classical approaches. This study shows preliminary results making such evaluations for the multisite precipitation synthesis task. We compared two open-source weather generators: IBMWeathergen (an extension of the Weathergen library) and RGeneratePrec, and two deep generative models: GAN and VAE, on a variety of metrics. Our preliminary results can serve as a guide for improving the design of deep learning architectures and algorithms for the multisite precipitation synthesis task.

\end{abstract}

\section{Introduction}

Stochastic weather generators are a common statistical downscaling tool that explicitly utilize the probabilistic nature of physical phenomena to model the marginal, temporal and sometimes spatial aspects of meteorological variables. They were first conceptualized by \cite{Richardson1981} and have since become widely used to produce long surrogate time series and downscale future climate projections for climate impact assessments (e.g. \citet{Kilsby2007}). They remain in wide use today (e.g., \citet{Vesely2019}).

Stochastic weather generation poses a number of unique challenges and have received recent attention from the machine learning community (e.g., \citet{li2021}, \citet{Puchko2020}). For example, the data being modeled can be highly-imbalanced, contain spatio-temporal dependencies and exhibit various anomalies -- e,g., extreme weather events -- exacerbated by anthropogenic climate change. 

Motivated by the absence of work comparing and evaluating stochastic and deep generative weather generators, we hereby perform a systematic evaluation of four weather generators for multisite precipitation synthesis: two open-source stochastic weather generators -- the IBMWeathergen (an extension of the weathergen library; ) and RGeneratePrec; and two deep generative models based on GAN and VAE architectures. The four weather generators are evaluated for Palghar, India which experiences heavy rainfall during the southwestern summer monsoons from July through September. This provides a challenging, highly-imbalanced precipitation dataset for synthetic generation. We used several metrics commonly used in literature to compare the empirical distribution of the simulations and observations and different patterns found in data like dry and well counts, dry and well spell lengths, total annual/monthly precipitation, and wet counts \cite{mehan2017comparative,tseng2020evaluation,mehrotra2006comparison,semenov1998comparison}.

\section{Data and Methods}
\subsection{Palghar Moonson Dataset}
Daily precipitation data for Palghar, India is from the Climate Hazards Group Infrared Precipitation with Stations v2.0 (CHIRPS) dataset. It contains global interpolated daily precipitation values at a spatial resolution of 0.05$^{\circ}$. We constructed a dataset for training the weather generators by gathering the daily precipitation data from CHIRPS from the period 01/01/1981 to 31/12/2009 within a bounding box corresponding to the latitude longitude pairs: 19$^{\circ}$ N, 72$^{\circ}$E and 20$^{\circ}$ N, 73$^{\circ}$E.
The bounding box contains 400 latitude and longitude pairs (sites) with precipitation values.

\subsection{Weather generators}
\subsubsection{IBMWeathergen}
We customized the weathergen singlesite library to perform multisite precipitation generation. Our implementation follows the methodology described in \cite{apipattanavis2007semiparametric} and includes an ARIMA forecasting component as in \cite{steinschneider2012semiparametric}. 
The occurrence model uses a first-order homogeneous Markov chain per month with three sequence states (dry, wet, and extreme). An ARIMA model captures the low-frequency trend of the interannual variability of the annual precipitation. For the precipitation model, IBMWeathergen uses a KNN 1-lag bootstrap resampler\footnote{1-lag refers to the resampling process being constrained to the sequence of two consecutive days given by the first-order Markov chain} and a KDE estimator. The model has extrapolation capabilities given by the ARIMA component and spatial coherence is guaranteed through the use of the resampling technique  \cite{apipattanavis2007semiparametric}.

\subsubsection{RGeneratePrec}
The RGeneratePrec model models temporal occurrence using a heterogeneous Markov chain per month with a probability transition matrix estimated through  Generalized linear models with the logit link function. The multisite precipitation occurrence follows Wilks' approach \cite{wilks1998multisite}, which estimates binary states of precipitation amounts for each site as a function of the probability integral transform of Gaussian random numbers constrained to the probability transition matrix of the temporal occurrence model.  The precipitation amount is generated for the corresponding states using a copula model based on a non-parametric distribution of the monthly observed samples \cite{cordano2016tools}.

\subsubsection{VAE \cite{vaes}}
We used an encoder that gets $32x32x32$ input data with two convolution blocks followed by a bottleneck dense layer and two dense layers for optimizing $\mu_x$ and $\sigma_x$ that hold the latent space  that is sampled to derive a normally distributed $z$.
We reduce the input dimension by four before submitting the outcome to the bottleneck dense layer using a down-sampling stage per convolutional block. We applied RELU after the convolutional and dense layers. Input  $z$ goes into the decoder and into a dense layer  to be reshaped into 256 activation maps of size $8x8x8$. These maps are inputs to consecutive transposed convolution layers that up-sampling the data up to the original  $32x32x32$ size. A final convolution using one filter is applied to get the  outcome. 

\subsubsection{GAN  \cite{goodfellow}}
We used similar architectures. The generator's encoder receives $32x32x32$ input data and applies two convolution blocks followed by a bottleneck dense layer. The decoder receives the encoder output and feeds it to a dense layer to be reshaped into 256 activation maps with a size of $8x8x8$. These maps serve as input to consecutive transposed convolution layers that up-sampling the data up to the original  $32x32x32$ size. The discriminator network uses an encoder architecture with a classification layer to implement the discrimination loss used to train the generator network.

\section{Preliminary results}
We use the IBMWeathergen and the RGeneratedPrec to 
generate 50 simulations for each of the 29 years of the dataset within the described bounding box.
For the VAE and the GAN models, we generated 32 representative days of the monsoon period for the bounding box in analysis\footnote{This approach of generating 32 days for representing the monsoon period was due to the scarcity of data for training this kind of model.} 
\begin{figure}[ht]
\vskip 0.2in
\begin{center}
\centerline{\includegraphics[width=2in]{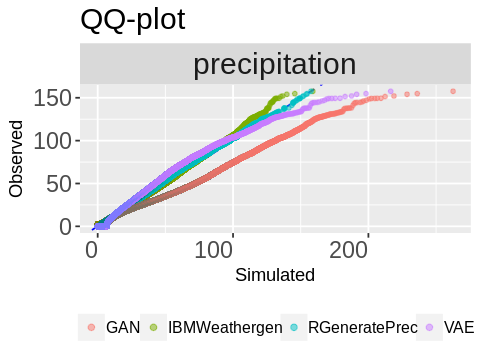}}
\caption{QQ-plot of observed vs. simulated precipitation values.}
\label{fig:global_qq_plot}
\end{center}
\vskip -0.2in
\end{figure}
Figure \ref{fig:global_qq_plot} shows a  comparison of the empirical distributions of observed and simulated values in terms of QQ-plots without considering the spatial locations and time of the year.
We observed that up to 100 mm/day, the IBMWeathergen and the RGeneratePrec models perform similarly. From the DL side, the VAE  follows the diagonal line closely, and the GAN fails to have a good representation of the distribution. Also, at dry observed days (0 mm/day) both VAE and GAN overestimate the wet days.
\begin{figure*}[ht]
\vskip 0.2in
\begin{center}
\centerline{\includegraphics[width=\textwidth]{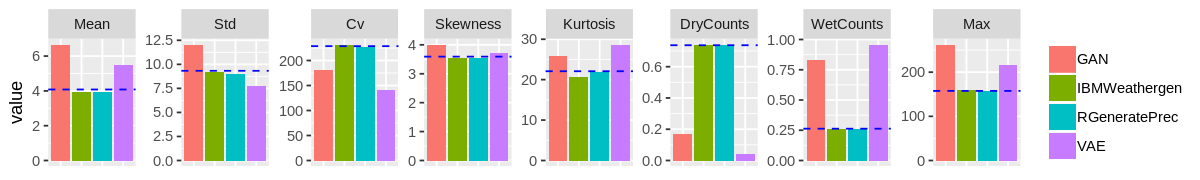}}
\caption{Comparison of observed and simulated precipitation values in terms of several quantitative measurements.}
\label{fig:global_stats}
\end{center}
\vskip -0.2in
\end{figure*}
We investigated the weather generators' simulated distribution in more detail as a function of several quantitative measurements without considering the spatial locations and time of the year. Figure \ref{fig:global_stats} shows this comparison in terms of the moments (mean, standard deviation\footnote{We are using the standard deviation instead of the variance because of intepretability}, skewness and kurtosis) and four quantitative measurements (coefficient of variation, wet counts, dry counts, and maximum values). In these results, the IBMWeathergen and the RGeneratePrec simulations represent the observed moments and quantitative measurements (dashed blue line). The GAN and the VAE models have a good approximation of the skewness, however they overestimate the mean, kurtosis, wet counts, and maximum values and underestimate the coefficient of variation and the dry counts.
\begin{figure*}[ht]
\vskip 0.2in
\begin{center}
\centerline{\includegraphics[width=\textwidth]{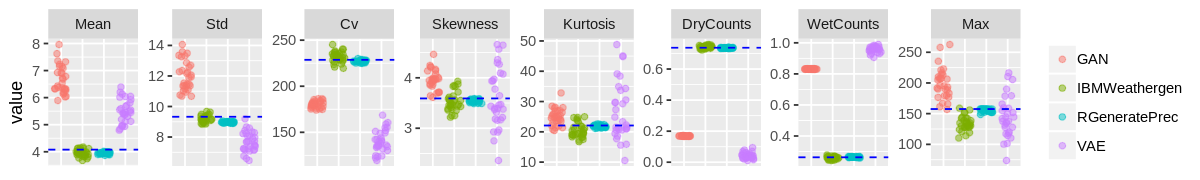}}
\caption{Comparison among empirical distributions of quantitative measures per simulation }
\label{fig:stats_per_simulation}
\end{center}
\vskip -0.2in
\end{figure*}
We performed the same analysis as above in the following experiment, although the moments and quantitative measurements were computed per simulation. Each point within the (Fig. \ref{fig:stats_per_simulation}) corresponds to a moment or quantitative measurement estimated from the precipitation values from individual simulations (without considering the spatial information and the time of occurrence).  The dashed blue line represents the quantitative measures of observed precipitation values. The results show that IBMWeathergen and RGeneratePrec have a better representation of those metrics than the DL models. The IBMWeathergen underestimates the maximum values, and it has more spread in representing the skewness and kurtosis than the RGeneratePrec.
On the other hand, the RGeneratePrec slightly underestimates the observed mean and standard deviation. GAN and VAE overestimate or underestimate all the metrics. VAE has a wider spread for skewness, kurtosis, and maximum values.
\begin{figure}[ht]
\vskip 0.2in
\begin{center}
\centerline{\includegraphics[width=2.5in]{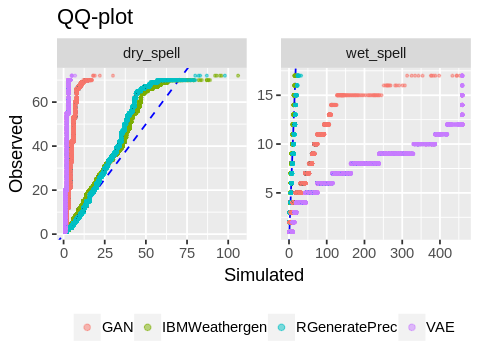}}
\caption{QQ-plots of dry and wet spell lengths of observed and simulated precipitation values.}
\label{fig:qq_plot_dry_wet_spell}
\end{center}
\vskip -0.2in
\end{figure}
Another experiment was to investigate if the weather generators could simulate the dry and wet spell length frequencies from the observed data. Figure \ref{fig:qq_plot_dry_wet_spell} shows this comparison in terms of QQ-plots. The results show that IBMWeathergen and RGeneratePrec can reproduce up to forty days of consecutive dry days found in the observed data. These two stochastic generators can also properly simulate the consecutive number of wet days found in the observations. On the other hand, GAN and VAE models fail to reproduce this information in the simulations.
\begin{figure}[ht]
\vskip 0.2in
\begin{center}
\centerline{\includegraphics[width=3in]{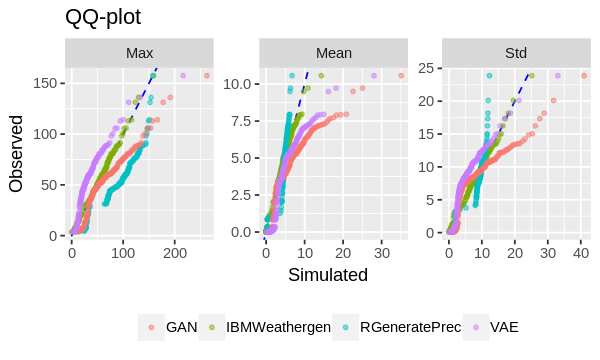}}
\caption{QQ-plots of the mean, standard deviation and maximum value per day: observed vs simulated.}
\label{fig:qq_day_year}
\end{center}
\vskip -0.2in
\end{figure}
One way to validate the simulations' temporal coherence is to analyze the simulated data at the day, month, and annual levels.  Figure \ref{fig:qq_day_year} shows a comparison of the distributions of the means, standard deviation, and maximum values per simulation day contrasted with the observed values. The results indicate that IBMWeathergen is better at representing those metrics followed by the VAE approach, while RGeneratePrec and GAN fail in simulating these metrics per day.
\begin{figure}[ht]
\vskip 0.2in
\begin{center}
\centerline{\includegraphics[width=\columnwidth]{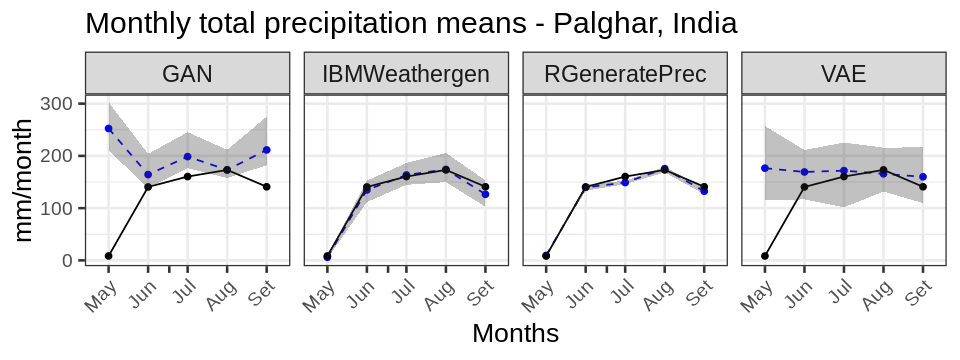}}
\caption{Means of the monthly total precipitation across the sites.}
\label{fig:monthly_total}
\end{center}
\vskip -0.2in
\end{figure}
\begin{figure}[ht]
\vskip 0.2in
\begin{center}
\centerline{\includegraphics[width=\columnwidth]{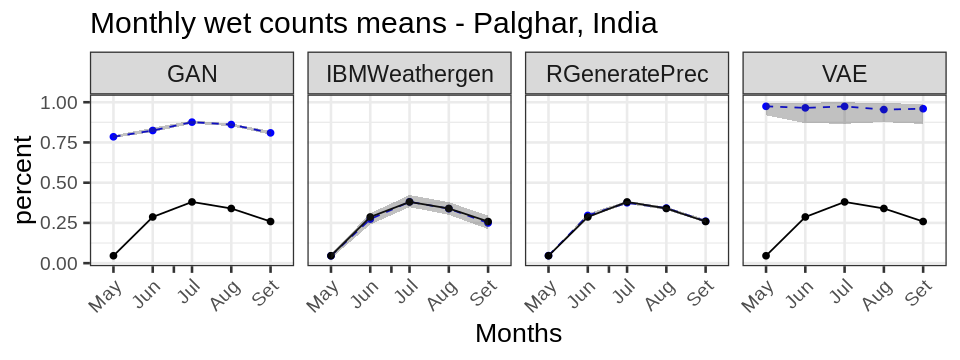}}
\caption{Means of the monthly wet counts across the sites.}
\label{fig:monthly_wet_counts}
\end{center}
\vskip -0.2in
\end{figure}
We explored the means of the monthly total precipitation and wet counts across the sites at the monthly level. Black points and lines in Figure \ref{fig:monthly_total} represent the means of the monthly total precipitation of observed values across the sites. Similarly, the blue points and lines are the medians of the monthly total simulated precipitation means. The limits of the gray area are the maximum and minimum of the monthly total simulated precipitation means. We observed that IBMWeathergen and RGeneratePrec simulations follow the observed monthly totals, with IBMWeathergen showing more variability. GAN overestimates the monthly total precipitation means. However, VAE shows promising results. It follows the monthly total precipitation means closely (except for May), and even it presents more variability, represented by a wider shade area, than the classical stochastic weather generators.
Figure \ref{fig:monthly_wet_counts} shows a similar experiment but in terms of percentage of the wet counts instead of precipitation. IBMWeathergen and RGeneratePrec successfully simulate this information whereas GAN and VAE overestimate the monthly wet counts.
\begin{figure}[ht]
\vskip 0.2in
\begin{center}
\centerline{\includegraphics[width=\columnwidth]{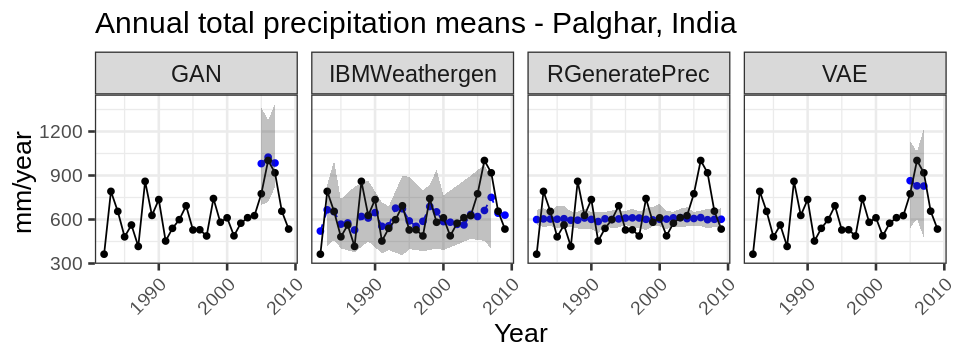}}
\caption{Means of the annual total precipitation across the sites.}
\label{fig:annual_total}
\end{center}
\vskip -0.2in
\end{figure}
\begin{figure}[ht]
\vskip 0.2in
\begin{center}
\centerline{\includegraphics[width=\columnwidth]{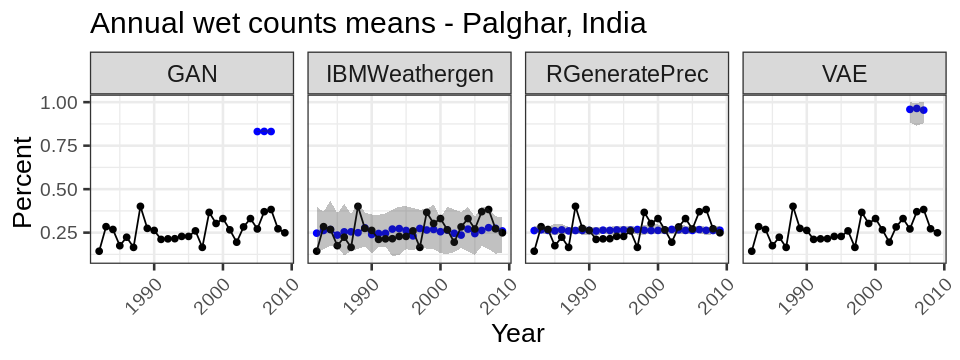}}
\caption{Means of the annual wet counts across the sites.}
\label{fig:annual_wet_counts}
\end{center}
\vskip -0.2in
\end{figure}
Finally, we explore whether or not the weather generators can reproduce the total annual precipitation and wet counts. Black points and lines in Fig. \ref{fig:annual_total} display the means of the total annual precipitation across all sites. The gray area identifies the limits of the means of the simulated total annual precipitation. Blue points and lines represent the medians, and the gray area limits are the maximum and minimum of the total annual simulated precipitation values across the sites. In this experiment, only IBMWeathergen can simulate the interannual variability while RGeneratePrec follows a linear trend pattern. As GAN and VAE models were not trained on specific years, they cannot distinguish the total annual variability. Figure \ref{fig:annual_wet_counts} shows the annual totals for GAN and VAE as reference, which overestimate the observed total annual precipitation. Figure \ref{fig:annual_wet_counts} shows a similar experiment but in terms of percentage of wet counts.  
\section{Discussion}
In this preliminary study, the IBMWeathergen model was consistently the best simulator for capturing different aspects of the observed precipitation values during the monsoon period in Palghar, India. However, there are other aspects we did not validate, including the superresolution capability of these generators for generating weather fields. (We hypothesize that the DL models can be better in this aspect, and we leave it as future research.) Deep learning applications in this realm are still immature. We hypothesize that it is possible to improve the design of weather generators based on deep learning methodologies by considering the metrics presented in this paper and others reported in the literature in the creation of loss functions, architectures, and algorithms\footnote{Research in stochastic weather generators is about 40 years old. The literature reports several methodologies for constructing them. However, there is still a lack of open source libraries and APIs ready for customization.}. For instance, open research questions are: How to constrain DL models to follow specific patterns found in data (e.g., dry/wet spell statistics)? How to couple DL models with temporal modeling concerning the annual and monthly variability? How to add control capability to deep learning models for generating extreme scenarios (extreme rainfalls, long dry/wet spells, etc.)? How to condition the models to forecasting values? and so on.






\bibliography{example_paper}

\begin{thebibliography}{15}
\providecommand{\natexlab}[1]{#1}
\providecommand{\url}[1]{\texttt{#1}}
\expandafter\ifx\csname urlstyle\endcsname\relax
  \providecommand{\doi}[1]{doi: #1}\else
  \providecommand{\doi}{doi: \begingroup \urlstyle{rm}\Url}\fi

\bibitem[Apipattanavis et~al.(2007)Apipattanavis, Podest{\'a}, Rajagopalan, and
  Katz]{apipattanavis2007semiparametric}
Apipattanavis, S., Podest{\'a}, G., Rajagopalan, B., and Katz, R.~W.
\newblock A semiparametric multivariate and multisite weather generator.
\newblock \emph{Water Resources Research}, 43\penalty0 (11), 2007.

\bibitem[Cordano et~al.(2016)Cordano, Eccel, et~al.]{cordano2016tools}
Cordano, E., Eccel, E., et~al.
\newblock Tools for stochastic weather series generation in r environment.
\newblock \emph{Ital J Agrometeorol}, 21:\penalty0 31--42, 2016.

\bibitem[Goodfellow et~al.(2014)Goodfellow, Pouget-Abadie, Mirza, Xu,
  Warde-Farley, Ozair, Courville, and Bengio]{goodfellow}
Goodfellow, I., Pouget-Abadie, J., Mirza, M., Xu, B., Warde-Farley, D., Ozair,
  S., Courville, A., and Bengio, Y.
\newblock Generative adversarial nets.
\newblock In Ghahramani, Z., Welling, M., Cortes, C., Lawrence, N., and
  Weinberger, K.~Q. (eds.), \emph{Advances in Neural Information Processing
  Systems}, volume~27. Curran Associates, Inc., 2014.
\newblock URL
  \url{https://proceedings.neurips.cc/paper/2014/file/5ca3e9b122f61f8f06494c97b1afccf3-Paper.pdf}.

\bibitem[Kilsby et~al.(2007)Kilsby, Jones, Burton, Ford, Fowler, Harpham,
  James, Smith, and Wilby]{Kilsby2007}
Kilsby, C., Jones, P., Burton, A., Ford, A., Fowler, H., Harpham, C., James,
  P., Smith, A., and Wilby, R.
\newblock A daily weather generator for use in climate change studies.
\newblock \emph{Environ. Model. Softw.}, 22:\penalty0 1705--1719, 2007.

\bibitem[Kingma \& Welling(2014)Kingma and Welling]{vaes}
Kingma, D.~P. and Welling, M.
\newblock {Auto-Encoding Variational Bayes}.
\newblock In \emph{2nd International Conference on Learning Representations,
  {ICLR} 2014, Banff, AB, Canada, April 14-16, 2014, Conference Track
  Proceedings}, 2014.

\bibitem[Li et~al.(2021)Li, Kou, and Zhao]{li2021}
Li, X., Kou, K., and Zhao, B.
\newblock Weather gan: Multi-domain weather translation using generative
  adversarial networks, 2021.

\bibitem[Mehan et~al.(2017)Mehan, Guo, Gitau, and
  Flanagan]{mehan2017comparative}
Mehan, S., Guo, T., Gitau, M.~W., and Flanagan, D.~C.
\newblock Comparative study of different stochastic weather generators for
  long-term climate data simulation.
\newblock \emph{Climate}, 5\penalty0 (2):\penalty0 26, 2017.

\bibitem[Mehrotra et~al.(2006)Mehrotra, Srikanthan, and
  Sharma]{mehrotra2006comparison}
Mehrotra, R., Srikanthan, R., and Sharma, A.
\newblock A comparison of three stochastic multi-site precipitation occurrence
  generators.
\newblock \emph{Journal of Hydrology}, 331\penalty0 (1-2):\penalty0 280--292,
  2006.

\bibitem[Puchko et~al.(2020)Puchko, Link, Hutchinson, Kravitz, and
  Snyder]{Puchko2020}
Puchko, A., Link, R., Hutchinson, B., Kravitz, B., and Snyder, A.
\newblock Deepclimgan: {A} high-resolution climate data generator.
\newblock \emph{CoRR}, abs/2011.11705, 2020.
\newblock URL \url{https://arxiv.org/abs/2011.11705}.

\bibitem[Richardson(1981)]{Richardson1981}
Richardson, C.~W.
\newblock Stochastic simulation of daily precipitation, temperature, and solar
  radiation.
\newblock \emph{Water Resources Research}, 17\penalty0 (1):\penalty0 182--190,
  1981.
\newblock \doi{https://doi.org/10.1029/WR017i001p00182}.

\bibitem[Semenov et~al.(1998)Semenov, Brooks, Barrow, and
  Richardson]{semenov1998comparison}
Semenov, M.~A., Brooks, R.~J., Barrow, E.~M., and Richardson, C.~W.
\newblock Comparison of the wgen and lars-wg stochastic weather generators for
  diverse climates.
\newblock \emph{Climate research}, 10\penalty0 (2):\penalty0 95--107, 1998.

\bibitem[Steinschneider \& Brown(2012)Steinschneider and
  Brown]{steinschneider2012semiparametric}
Steinschneider, S. and Brown, C.
\newblock A semiparametric multivariate and multi-site weather generator with a
  low-frequency variability component for use in bottom-up, risk-based climate
  change assessments.
\newblock In \emph{AGU Fall Meeting Abstracts}, volume 2012, pp.\  GC41B--0973,
  2012.

\bibitem[Tseng et~al.(2020)Tseng, Chen, and Senarath]{tseng2020evaluation}
Tseng, S.-C., Chen, C.-J., and Senarath, S.~U.
\newblock Evaluation of multi-site precipitation generators across scales.
\newblock \emph{International Journal of Climatology}, 40\penalty0
  (10):\penalty0 4622--4637, 2020.

\bibitem[Vesely et~al.(2019)]{Vesely2019}
Vesely, F. et~al.
\newblock Quantifying uncertainty due to stochastic weather generators in
  climate change impact studies.
\newblock \emph{Sci. Rep.}, 9:\penalty0 9258, 2019.

\bibitem[Wilks(1998)]{wilks1998multisite}
Wilks, D.
\newblock Multisite generalization of a daily stochastic precipitation
  generation model.
\newblock \emph{journal of Hydrology}, 210\penalty0 (1-4):\penalty0 178--191,
  1998.

\end{thebibliography}
\bibliographystyle{icml2021}



\end{document}